\documentclass[10pt,twocolumn,letterpaper]{article}

\usepackage{cvpr}              

\definecolor{cvprblue}{rgb}{0.21,0.49,0.74}
\usepackage[pagebackref,breaklinks,colorlinks,allcolors=cvprblue]{hyperref}

\usepackage{stfloats}
\usepackage{booktabs}
\usepackage{framed,multirow}
\usepackage{array}
\usepackage{colortbl}
\newlength\savewidth\newcommand\shline{\noalign{\global\savewidth\arrayrulewidth
  \global\arrayrulewidth 1pt}\hline\noalign{\global\arrayrulewidth\savewidth}}
\newcommand{\tablestyle}[2]{\setlength{\tabcolsep}{#1}\renewcommand{\arraystretch}{#2}\centering\footnotesize}

\newcolumntype{x}[1]{>{\centering\arraybackslash}p{#1pt}}
\newcolumntype{y}[1]{>{\raggedright\arraybackslash}p{#1pt}}
\newcolumntype{z}[1]{>{\raggedleft\arraybackslash}p{#1pt}}
\newcommand{\app}{\raise.17ex\hbox{$\scriptstyle\sim$}}

\definecolor{baselinecolor}{gray}{.9}

\newcolumntype{*}{>{\global\let\currentrowstyle\relax}}
\newcolumntype{^}{>{\currentrowstyle}}

\definecolor{dt}{gray}{0.7}  %

\def\paperID{8019} 
\def\confName{CVPR}
\def\confYear{2025}

\title{Progressive Vision-Language Prompt for Multi-Organ Multi-Class Cell Semantic Segmentation with Single Branch}

\author{Qing Zhang$^{1}$, Hang Guo$^{1}$, Siyuan Yang$^{2}$, Qingli Li$^{1}$, Yan Wang$^{1}$\thanks{Corresponding author.} 
\\
$^1$ Shanghai Key Laboratory of Multidimensional Information Processing, \\East China Normal University, Shanghai, China.\\
$^2$ The School of Electrical and Electronic Engineering, Nanyang Technological University, Singapore.\\
e-mail: qzhang@ce.ecnu.edu.cn}

\begin{document}
\maketitle
\begin{abstract}
Pathological cell semantic segmentation is a fundamental technology in computational pathology, essential for applications like cancer diagnosis and effective treatment.
Given that multiple cell types exist across various organs, with subtle differences in cell size and shape, multi-organ, multi-class cell segmentation is particularly challenging.
Most existing methods employ multi-branch frameworks to enhance feature extraction, but often result in complex architectures.
Moreover, reliance on visual information limits performance in multi-class analysis due to intricate textural details.
To address these challenges, we propose a Multi-OrgaN multi-Class cell semantic segmentation method with a single brancH (MONCH) that leverages vision-language input.
Specifically, we design a hierarchical feature extraction mechanism to provide coarse-to-fine-grained features for segmenting cells of various shapes, including high-frequency, convolutional, and topological features.
Inspired by the synergy of textual and multi-grained visual features, we introduce a progressive prompt decoder to harmonize multimodal information, integrating features from fine to coarse granularity for better context capture.
Extensive experiments on the PanNuke dataset, which has significant class imbalance and subtle cell size and shape variations, demonstrate that MONCH outperforms state-of-the-art cell segmentation methods and vision-language models.
Codes and implementations will be made publicly available.
\end{abstract}
\vspace{-3mm}
\section{Introduction}
\label{sec:intro}
Multi-organ, multi-class, multi-cell segmentation and classification involves segmenting cell contours for different cell types in digitized patient specimens, such as Whole Slide Images (WSIs) \cite{oh2024controllable,verma2021monusac2020,fehri2019bayesian,meng2018large}.
It is a foundational technology in computational pathology, enabling both quantification and visualization of various cell types to provide a reliable basis for diagnosis in clinical applications such as prognosis evaluation, cancer grading, and treatment planning \cite{kassim2020clustering,punn2020inception,petukhov2022cell}.
Accurate multi-organ, multi-cell segmentation models are crucial in determining the nature, grade, and stage of diseases, making them highly valuable for clinical practice and worthy of further exploration.

\vspace{-1mm}
Deep learning has rapidly advanced, achieving impressive performance in nuclei instance segmentation \cite{xie2020instance,he2023toposeg} and multi-class cell semantic segmentation \cite{bokhorst2023deep,berger2024topologically}. 
Despite recent progress, comprehensively analyzing multi-class cells from various organs remains challenging due to the difficulty in capturing accurate features of different cell types across organs. 
Additionally, the high imbalance in these datasets presents another challenge for the model's feature extraction capabilities.
%
Learning from less-diverse data makes it difficult for these models to transfer to other organs, as training datasets do not match the data distribution found in `the clinical wild' \cite{gamper2019pannuke,gamper2020pannuke}.
One solution to better extract pathological features across organs is to split the cell segmentation task into multiple branches, such as nuclei instance segmentation, position prediction, and classification~\cite{graham2019hover,dogar2023attention}. 
While these multi-branch methods have shown good performance in multi-organ multi-class cell segmentation, their model complexity and efficiency are significantly lower than those of single-branch methods. 
In physiological environments, various cell types coexist, each exhibiting distinct textures and sizes.
How to fully extract latent information and strong semantic features from feature maps is crucial for the downstream tasks. 
However, few of these methods consider multi-grained features.  Existing feature fusion methods simply concatenate multi-level features, which often leads to redundant information.
To address this, feature pyramid and attention-based methods have been proposed for effective multi-scale feature fusion \cite{guo2020augfpn,chen2021crossvit,xu2023collaborative}.
Feature Pyramid Networks (FPNs), however, may lose important information during pooling or downsampling operations, and their performance can degrade when applied to datasets with significantly different distributions.
Attention-based methods enable information transfer between multi-grained feature maps, enhancing feature representation by capturing more comprehensive contextual information \cite{liu2021learning,wu2023aggn}.

As mentioned above, traditional single-branch methods also struggle to capture diverse features, as they rely purely on semantic information, limiting performance when data exhibits significant diversity.
Vision-Language Models (VLMs) integrate computer vision and natural language processing, enhancing semantic features by incorporating textural information \cite{xu2022simple,xu2023learning,gao2024enhancing}.
VLMs offer strong feature extraction capabilities due to pre-training on large-scale datasets.
Their ability to align image and text features enables the use of textual information to supplement image segmentation.
Despite their successes, the unified potential of VLMs for pathological cell segmentation remains largely underexplored.

To address these issues, we propose a network for Multi-OrgaN multi-class multi-Cell segmentation with a single-brancH, named MONCH, which combines progressive prompts with textual attributes and multi-grained visual features.
First, textual attributes of different cell types are generated using GPT-4, followed by visual-textual feature fusion.
Given the varying shapes and sizes of different cell types, a multi-grained visual feature extraction block (MGFE) is designed to extract comprehensive visual features, including high-frequency details, semantic information, and topological features capturing mutual information among multiple cells.
To fully integrate textual and multi-grained visual information, we introduce a progressive prompt decoder (PPD) that gradually merging features from fine-grained visual feature to coarse-grained textual feature, applying the finer feature as the query for the coarse feature. 


The main contributions of this work are as follows:
\begin{itemize}
    \item We propose MONCH, a single-branch network leveraging textual and visual information, which effectively segments multiple cell types across various organs through comprehensive feature extraction and a progressive prompt decoder.
    \item We design the MGFE block to extract multi-grained features from image features enhanced by a pre-trained VLM, enabling MONCH to capture detailed visual information.
    \item We introduce the PPD block to integrate features from fine to coarse granularity, aiding in capturing global context while preserving visual details.
    \item Extensive experiments conducted on the public PanNuke dataset demonstrate that MONCH achieves state-of-the-art performance in multi-organ, multi-class, multi-cell segmentation.
\end{itemize}

\section{Related Work}
\label{sec:relatedwork}
\subsection{Multi-Class Cell Detection and Segmentation}
The distribution of multi-class pathological cells provides crucial auxiliary diagnostic information for pathologists.
A common approach for analyzing multi-class cell distribution is to divide the cell segmentation framework into multiple branches, such as performing object segmentation followed by classification \cite{abousamra2021multi,xie2024db}.
This approach helps eliminate background interference from original pathological images, leading to improved cell classification outcomes.
Some research further adds branches to enhance contour or texture information of different cell types \cite{he2023transnuseg,graham2019hover}.
For instance, Meta-MTL proposes a multi-task nuclei segmentation network with both contour detection and segmentation tasks, with a feature attention module to amplify shape information \cite{han2022meta}.
Similarly, AL-Net~\cite{zhao2021net} introduces an attention-based learning network using multi-task learning strategy to enhance segmentation feature extraction by predicting nuclei boundaries.
Beyond boundary detection, several studies focus on enhancing textual feature learning through semantic information analysis.
For example, SMILE~\cite{pan2023smile} leverages cell semantic segmentation and instantiation to generate a distance transformation map, improving the accuracy of multi-class cell instance segmentation.
GSN-HVNET~\cite{zhao2023gsn} employs an encoder-decoder framework to extract precise cell features for segmentation and classification tasks.

While multi-branch cell segmentation networks have shown strong performance in multi-class cell segmentation and classification, their increased computational complexity is a significant disadvantage.
Therefore, designing a single-branch network that efficiently merges diverse features to achieve accurate multi-class cell segmentation remains an important research question.

\begin{figure*}[t]
\centering
    \includegraphics[width=0.97\textwidth]{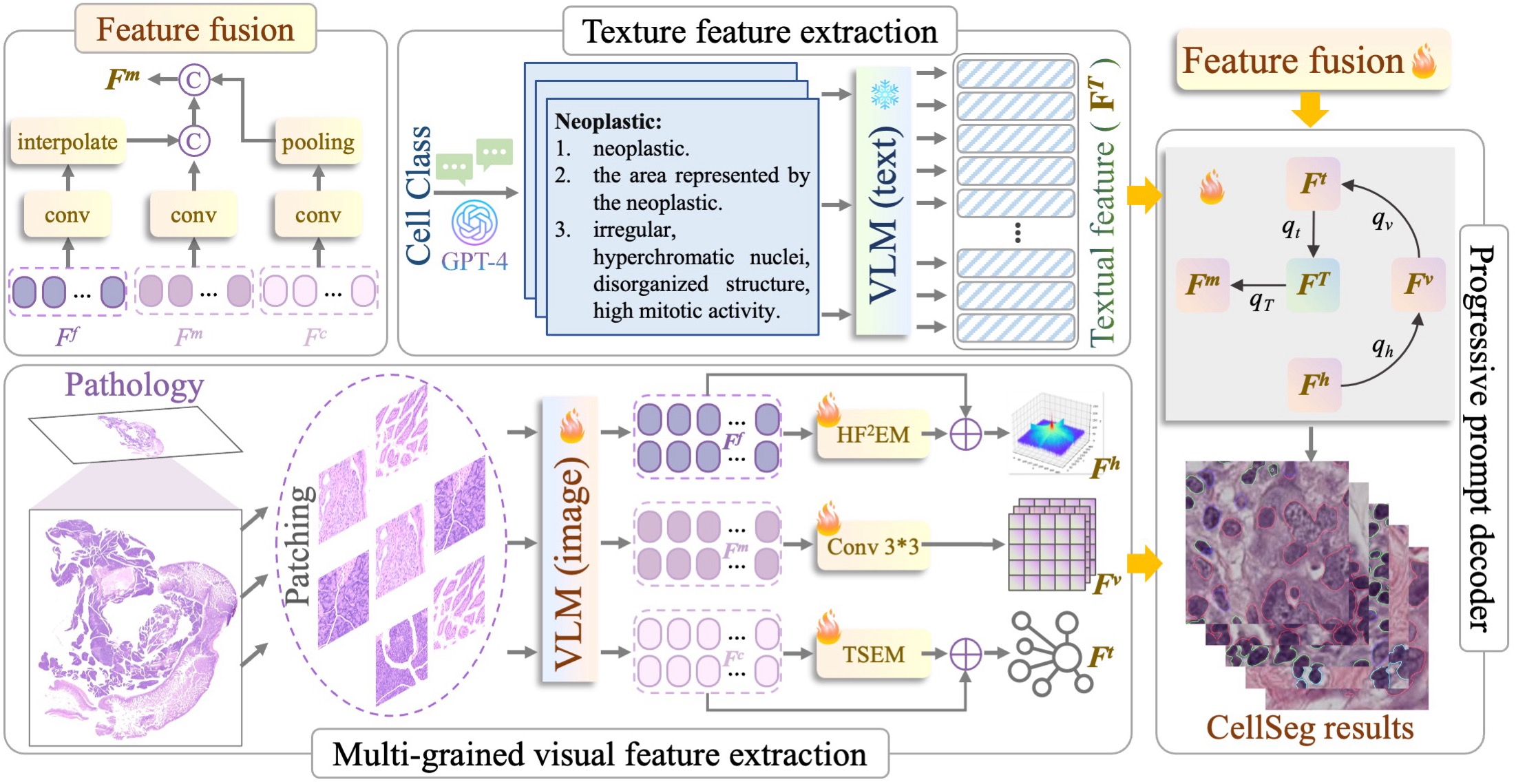}
    \caption{Overview of the proposed method. \emph{Texture feature extraction}: Textual features are extracted via a frozen text encoder based on GPT-generated cell attributes. \emph{Multi-grained visual feature extraction}: Multi-grained visual features are obtained from a pre-trained image encoder and enhanced via specific feature extraction modules. $HF^2EM$ is a high-frequency extraction module, $Conv 3*3$ is a convolutional block, and $TSEM$ is a topological structure extraction module. \emph{Feature fusion}: Multi-scale visual features are integrated using feature pyramid fusion block. \emph{Progressive prompt decoder}: Multimodal features are progressively input into the cross-attention module as prompts to lower-level features, harmonizing the discrepancy between multi-grained visual and linguistic features.}
    \label{fig: framework}
\end{figure*}

\subsection{Multi-Scale Feature Learning}
Given the diverse structural and scalar details present across multiple cell types, multi-scale feature fusion is crucial for capturing information ranging from low-level textures and edges to high-level semantics and contextual details.
Deep learning networks extensively used for image fusion are generally categorized into CNN-based and attention-based architectures.
CNN-based methods leverage their architectural design to integrate features across various scales, e.g., FPN \cite{guo2020augfpn,ju2021adaptive} and skip connections \cite{wang2022uctransnet,yan2020attention}, etc.
However, CNN-based feature fusion methods typically use static input features, which increases computational demands.
Attention-based methods, on the other hand, provide nonlinear strategies for multi-scale feature fusion, effectively handling features of varying semantics and scales.
These methods can adjust input features dynamically, leading to the proliferation of attention-based fusion networks \cite{tong2023msaffnet,wu2023aggn}.
For example, MS-CAM~\cite{dai2021attentional} proposes a multi-scale channel attention module for feature fusion.
HiFuse~\cite{huo2024hifuse} introduces a three-branch hierarchical medical image classification network with a self-attention module to integrate local multi-scale features.
In the field of medical image analysis, several attention-based multi-scale feature fusion networks have been proposed, including AF-Net\cite{hou2021af}, MAXFormer \cite{liang2023maxformer} and MSA-Net \cite{wang2024msa}, etc.
While these attention-based methods generally outperform CNN-based approaches in multi-scale feature integration, current cross-attention mechanisms still struggle to fully capture the complex relationships between features at different scales.

\subsection{Vision-Language Based Fine-Tuning}
Most visual analysis research focuses on training a single-branch deep neural network using extensive annotated datasets, resulting in a laborious and time-consuming process \cite{minaee2021image,yu2023techniques}.
Recently, Vision-Language Models (VLMs) have gained significant attention for their ability to learn intricate correlations between visual and linguistic features using web-scale image-text pairs \cite{zhang2024vision,zhou2022learning}. Leveraging VLMs has had a substantial impact on computational pathology.
To be more specific, VLMs have become particularly popular in computational pathology since the introduction of Contrastive Language-Image Pre-training (CLIP) \cite{radford2021learning}.
For example, PLIP~\cite{huang2023visual} introduces a novel approach to pathology by developing a language-image pre-training model capable of analyzing multimodal data.
UNI~\cite{chen2024towards} effectively adapts VLMs originally trained on natural images to various pathological tasks by leveraging large-scale pathological image-text pairs.
CONCH~\cite{lu2024visual} proposes a pioneering vision-language foundation model that employs contrastive learning on image-caption pairs, addressing several downstream tasks such as image analysis, text-to-image, and image-to-text retrieval.

\vspace{-1mm}
While vision-language foundation models have shown promising results in various tasks, they primarily focus on organ-level analysis.
However, subtle changes and interactions within cells can provide pathologists with more detailed structural and functional insights, allowing for the identification and examination of heterogeneity among cellular populations \cite{hao2024large}.
Therefore, further exploration of cell-level VLMs is a promising area of research.

\section{Method}
\label{sec:method}
\subsection{Problem Setting and Network Architecture}
Mathematically, given a WSI $\mathbf{X}_w\in\mathbb{R}^{W_w\times H_w\times 3}$, whose size is $W_w\times H_w$, a set of patches $\mathbf{X}\in\mathbb{R}^{W_p\times H_p\times 3}$ are cropped from the WSI.
These images contain $C$ different cell types and are collected from multiple organs.
Our goal is to predict the pixel-level label $\hat{\mathbf{Y}}=\{0,1,...,N\}\in\mathbb{R}^{W\times H}$, where $N$ is the number of cell types, based on both image and textual prompts.

To address this problem, we propose a novel framework for Multi-OrgaN multi-class multi-Cell segmentation with a single-brancH, named MONCH.
As shown in Fig.~\ref{fig: framework}, the proposed MONCH incorporates a coarse-to-fine visual feature extraction mechanism with a progressive vision language prompt decoder to efficiently fuse textural and multi-grained visual features.
Specifically, we first generate cell attributes $\mathcal{T}$ utilizing GPT-4, which encompasses the background description and $N$ cell type descriptions. Among $N+1$ types, we describe each of them using three sentences. 
We then encode cell descriptions $\mathbf{S}$ and pathological cell image $\mathbf{X}_p$ using the image and text encoders inherited from a pre-trained VLM, thereby obtaining the textual feature $\mathbf{F}^\mathcal{T}$ and multi-grained image features $\mathbf{F_X}$. $\mathbf{F^X}$ is calculated as:
\begin{equation}
\small
\mathbf{F^X} = \{\mathbf{F}^c, \mathbf{F}^m, \mathbf{F}^f\}, 
\end{equation}
where $\mathbf{F}^m = \mathcal{G}_{VLM}(\mathbf{X}, \mathbf{S})$, $\mathbf{F}^c = \mathcal{G}_{ds}(\mathbf{F}^m)$, and $\mathbf{F}^f = \mathcal{G}_{us}(\mathbf{F}^m)$. Here, $\mathcal{G}_{VLM}(\cdot)$ is the image encoder of the pre-trained VLM, $\mathcal{G}_{ds}(\cdot)$ is the downscale block, and $\mathcal{G}_{us}(\cdot)$ is the upscale block.
The generated multi-grained features are middle-grained $\mathbf{F}^m\in\mathbb{R}^{B\times C\times W\times H}$, fine-grained $\mathbf{F}^f\in\mathbb{R}^{B\times C\times 2W\times 2H}$, and coarse-grained $\mathbf{F}^c\in\mathbb{R}^{B\times C\times W/2\times H/2}$, where $B$ is the batch size and $C$ is the channel size.
Given the varying textures and scale information across different cell types, we introduce a coarse-to-fine feature extraction module to enhance the multi-grained features generated from the pre-trained image encoder.

Fine-grained features capture intricate details of pathological cells, which motivates the integration of a high-frequency information extraction module to enhance cell textual features.
Coarse-grained features, while lacking in detailed information, retain essential cell distribution insights.
To further enrich the structural and semantic content, we propose learning topological features.
To preserve the original characteristics of the image, we incorporate a convolutional block to capture local and textual nuances. 

The aforementioned comprehensive visual features, raging from coarse- to fine-grained, are then integrated with the embedded features derived from attribute prompts, resulting in enhanced image features that simultaneously capture visual and textural attributes of cells.
To fully integrate these multimodal features, we design a progressive vision-language prompt decoder that iteratively adopts one feature set as the query for the subsequent feature set at a finer level.
The coarse features are refined through fusing them with the multi-head self-attention mechanism, utilizing fine-grained features as the guiding query.
Based on this progressive prompt learning approach, all features are effectively integrated, thereby facilitating superior performance in multi-class cell semantic segmentation.

\begin{figure}[t]
    \centering
    \includegraphics[width=\linewidth]{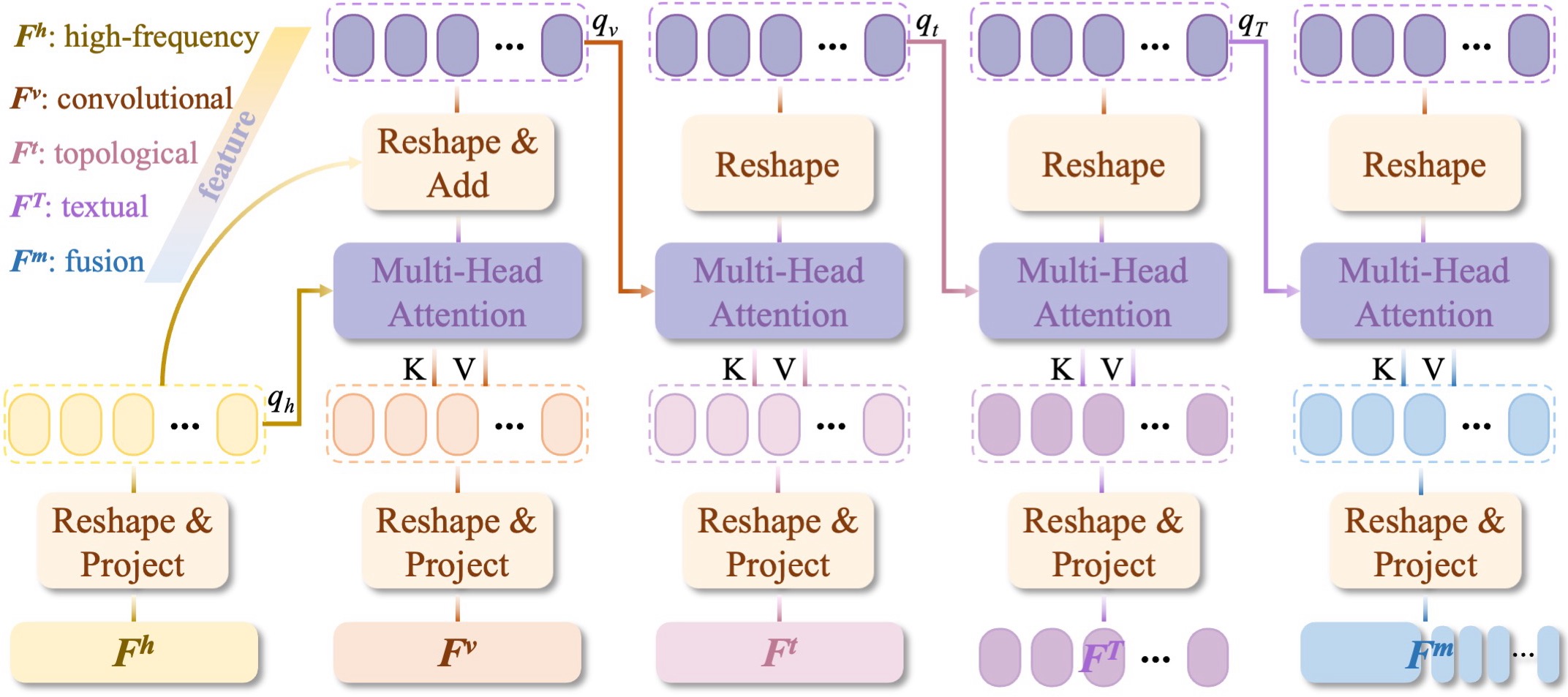}
    \vspace{-1mm}
    \caption{Progressive Vision-Language Prompt Decoder: Multimodal information, including textual features and multi-grained visual features, progressively serve as queries in a multi-head self-attention to harmonize features from fine-coarse-fine granularity.
    }
    \vspace{-5mm}
    \label{fig:ppd}
\end{figure}

\subsection{Coarse-to-Fine Visual Feature Enhancement}
To enhance the representation of pathological cells, we introduce a coarse-to-fine visual feature extraction module for multi-grained visual feature enhancement.
Our approach utilizes text and image encoders from a pre-trained VLM to align textual and visual features.
The initial medium-grained image feature, $\mathbf{F}^m$, is obtained from the pre-trained image encoder, capturing fundamental local details such as cell edges, textures, and essential structures
%
%
We further process $\mathbf{F}^m$ by upscaling it to obtain a fine-grained feature $\mathbf{F}^f$ and downscaling it to derive a coarse-grained feature $\mathbf{F}^c$.
These transformations provide multi-grained perspectives of the pathological cells.
A convolutional module is then applied to adapt $\mathbf{F}^m$ for improved suitability to the new task, resulting in a convolutional feature $\mathbf{F}^v = \text{conv}(\mathbf{F}^m)$.
For fine-grained feature enhancement, we design a high-pass filter module that produces a refined high-frequency image feature $\mathbf{F}^h$. This process begins by transforming the fine-grained feature $\mathbf{F}^f$ at position $(x,y)$ into the frequency domain using a Fourier Transform $\mathcal{F}(\cdot)$:
\begin{equation}
\small
    \mathbf{F}^{f'}(x,y)=\mathcal{F}(\mathbf{F}^f(x,y)).
\end{equation}
We then extract the high-frequency feature by applying a high-pass filter $\mathcal{H}(\cdot)$, based on the resolution of $\mathbf{F}^f$:
\begin{equation}
\small
    \mathbf{F}^{f''}(x,y)=\mathcal{H}(\mathbf{F}^{f'}(x,y)),
\end{equation}
where $\mathbf{F}^{f''}$ represents the high-frequency component. The refined high-frequency feature $\mathbf{F}^h$ is obtained through an inverse Fourier transform $\mathcal{F}^{-1}(\cdot)$ and a residual operation:
\begin{equation}
\small
    \mathbf{F}^h(x,y)=\mathcal{F}^{-1}(\mathbf{F}^{f''}(x,y))+\mathbf{F}^f(x,y).
\end{equation}

For the coarse-grained visual feature, we introduce a topological feature extraction module to better capture the intrinsic structural and shape information.
Given the coarse-grained feature $\mathbf{F}^c$, we apply dilated KNN to calculate pixel-level pairwise distances, obtaining $k=9$ nearest neighbors for each point with dimensions $B\times C/2\times k\times k$.
We first normalize $\mathbf{F}^c$ to reduce its channel dimension to get a blended feature with dimension of $B\times C/2\times W/2\times H/2$.
For each image feature $\mathbf{F}^{c'}$ in its spatial dimension, the pixel-level pairwise distance is calculated as:
\begin{equation}
\small
    \mathbf{D} = ||{\mathbf{F}^{c'}}||^2,
    \hat{\mathbf{D}} = \mathbf{D}-2\times(\mathbf{F}^{c'}\cdot{\mathbf{F}^{c'}}^\top)+\mathbf{D}^\top,
\end{equation}
where $\mathbf{D}$ is the pairwise distance of $\mathbf{F}^{c'}$.
The $k$ nearest neighbors for each spatial-level feature are then extracted based on the distance $\hat{\mathbf{D}}$, resulting in the topological structure $\mathbf{F}^k \in \mathbf{R}^{B\times C/2\times k\times \times k}$. 
The enhanced topological feature $\mathbf{F}^t$ is obtained through a residual connection:
\begin{equation}
\small
    \mathbf{F}^t=\mathbf{F}^k+\mathbf{F}^c.
\end{equation}

\noindent By combining these enhanced multi-grained visual features, the proposed MONCH comprehensively captures both local fine details and global structural information, leading to a more complete representation of pathological cells.

\begin{figure*}[t]
    \centering
    \includegraphics[width=0.97\linewidth]{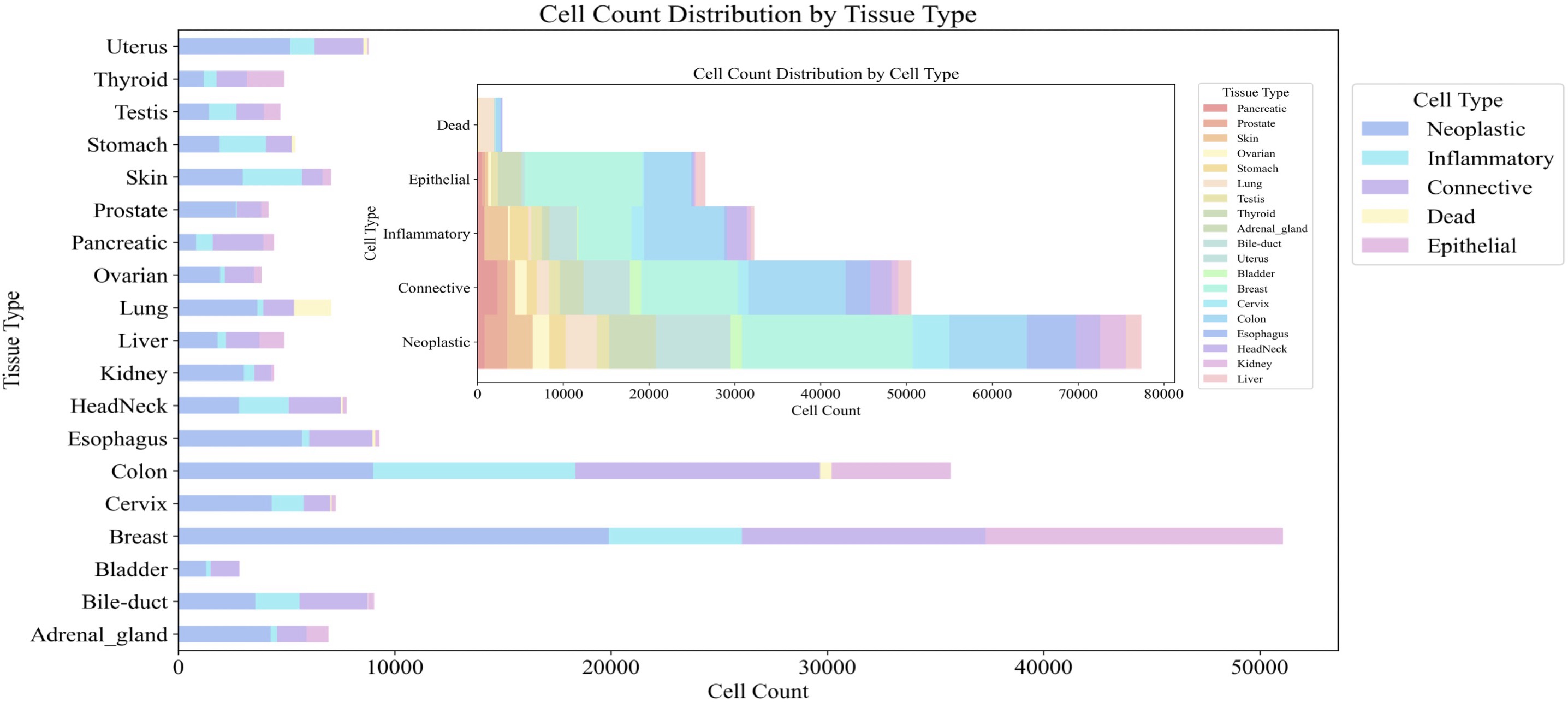}
    \caption{PanNuke Cell Distribution Map. Distribution of each of the 19 organ types and 5 cell types.}
    \vspace{-2mm}
    \label{fig:datadistribution}
\end{figure*}
\subsection{Progressive Prompt Decoder}
To effectively harmonize the linguistic features with the multi-grained visual features, we introduce a progressive vision-language prompt decoder that generates a comprehensive representation for enhanced analysis, as illustrated in Fig.~\ref{fig:ppd}.
In this framework, the higher-level features are sequentially used as queries for lower-level features to get a robust representation.

To reconcile knowledge differences between these modalities and extract additional textual information, we propose an attention mechanism that leverages the complementarity of visual and language features.
Finer $query$ implies higher distinctiveness in attention mechanism, resulting in better differentiating between various parts of the input feature and capturing meaningful contextual information.
Therefore, this attention based hierarchical approach ensures that the $query$ feature effectively captures the core requirements of the task, providing key details for refinement at each level.
MONCH progressively sets the fine-grained feature as the $query$ for the coarser-grained feature from $\mathbf{F}^h$ to $\mathbf{F}^\mathcal{T}$, i.e. $\mathbf{F}^h \to \mathbf{F}^v \to \mathbf{F}^t \to \mathbf{F}^\mathcal{T}$.
Specifically, using \(\mathbf{F}^h\) as the $query$ for \(\mathbf{F}^v\), the multi-head self-attention module is calculated as follows:
\begin{equation}
\small
\label{ms}
\begin{aligned}
    \mathcal{G}_{ms}(\mathbf{F}^h, \mathbf{F}^v)=\mathtt{softmax}(\frac{g_q(\mathbf{F}^h) \cdot {g_k(\mathbf{F}^v)}^\top}{\sqrt{d_k}}) \cdot g_v(\mathbf{F}^v),
\end{aligned}
\end{equation}
where $\mathcal{G}_{ms}(\cdot)$ represents the multi-head self-attention layer, \(g_q(\cdot)\), \(g_k(\cdot)\) and \(g_v(\cdot)\) are projection functions, and \(d_k\) is the dimension of the \(key\).
The subsequent multi-head self-attention modules are calculated in the same manner.

As shown in Fig.~\ref{fig:ppd}, after three-iteration attention architecture progressing from fine-grained to coarse-grained attention, The interactively learned features are ultimately merged with a blended feature, which is generated by fusing the three-scale features $\{\mathbf{F}^c, \mathbf{F}^m, \mathbf{F}^f\}$ obtained from the pre-trained image encoder using an FPN-like feature fusion block in Fig.~\ref{fig: framework}.
To enhance the detailed cell feature, we set the iteratively generated visual feature as the $query$ to this blended feature. 
Through this progressive prompt decoder strategy, we effectively harmonize multimodal and multi-granular features of pathological cells, ensuring accurate integration and representation of diverse characteristics.
A final merge step is designed to enhance feature richness, resulting in a more robust representation.

\subsection{Loss Function}
The proposed MONCH builds upon a pre-trained VLM architecture, enhanced with several specifically designed adapters.
Our network takes as input a combination of limited pathological cell descriptions and their associated images.
To optimize learning, we freeze the text encoder while fine-tuning the image encoder, allowing better adaptation to pathological cell feature analysis.
Following the progressive vision-language prompt decoder, we obtain a merged feature $\mathbf{F}^m$ that effectively integrates both textual and visual features.
To further refine this representation, we introduce a forward propagation module that takes the textual feature $\mathbf{F}_\mathcal{T}$ and the merged feature $\mathbf{F}^m$ as inputs:
\begin{equation}
\small
    \mathbf{F}^m=\mathcal{G}_{conv}(\mathcal{G}_{reshape}( \mathcal{G}_v(\mathbf{F}^m),\mathcal{G}_{weight}(\mathcal{G}_{linear}(\mathbf{F}_\mathcal{T})))),
\end{equation}
where $\mathcal{G}_{reshape}$ is operation that transforms a feature with a resolution of $B\times C\times W\times H$ into a matrix with dimensions $(1, B\times C, W, H)$. $\mathcal{G}_v$ represents a series of sequential convolution layers for processing visual features. $\mathcal{G}_{linear}$ is a linear transformation applied to the textual feature, and $\mathcal{G}_{weight}$ reshapes the weights for the subsequent convolution operation $\mathcal{G}_{conv}$.

For accurate multi-class cell semantic segmentation, we introduce a segmentation loss function $\mathcal{L}_{seg}$. We employ binary cross-entropy loss to facilitate precise segmentation of each cell class:
\begin{equation}
\small
    \mathcal{L}_{seg}(y,p)=
    -\frac{1}{N} \frac{1}{B} \sum_{n=1}^N \sum_{i=1}^B[y_{in}\log(p_{in})+(1-y_{in})\log(1-p_{in})],
\end{equation}
where $y_{in}$ is the $i^{th}$ ground truth for cell class $n$, and $p_{in}$ is the $i^{th}$ predicted segmentation map for $n^{th}$ cell class.
\section{Results}
\label{sec: results}
\begin{table*}[ht]
    \tablestyle{6pt}{1}
    \caption{Evaluation against SOTA cell segmentation methods in cell types from PanNuke. The best results are highlighted in \textbf{bold}.}
    \resizebox{\textwidth}{!}{
    \begin{tabular}{*l ^c ^c ^c  ^c ^c ^c  ^c ^c ^c  ^c ^c ^c  ^c ^c ^c ^c}
        \shline
        \multirow{2}{*}{Model}   &\multicolumn{3}{c}{Neoplastic} &\multicolumn{3}{c}{Epithelial} &\multicolumn{3}{c}{Inflammatory} &\multicolumn{3}{c}{Connective} &\multicolumn{3}{c}{Dead} &\multirow{2}{*}{Inference (ms)}\\
        \cmidrule(r){2-4} \cmidrule(r){5-7} \cmidrule(r){8-10} \cmidrule(r){11-13} \cmidrule(r){14-16} 
        ~ &$IoU$ &$Precision$ &$F1 Score$  &$IoU$ &$Precision$ &$F1 Score$  &$IoU$ &$Precision$ &$F1 Score$  &$IoU$ &$Precision$ &$F1 Score$  &$IoU$ &$Precision$ &$F1 Score$\\
        \shline
        HoVer-Net \cite{graham2019hover}     &0.6343 &0.7372 &0.7762    &0.5171 &0.6999 &0.6817    &0.4515 &\textbf{0.7188} &0.6221    &0.4316 &0.6496 &0.6030    &0.1479 &0.2130 &0.2577 &98.218\\
        TSFD-Net \cite{ilyas2022tsfd}      &0.6573 &0.7418 &0.7932    &0.5893 &0.7195 &0.7416    &\textbf{0.5153}&0.6983&\textbf{0.6801}    &0.4600 &0.6649 &0.6301    &0.0118 &0.4215 &0.0234 &742.81\\
        CPP-Net \cite{chen2023cpp}       &0.6497 &\textbf{0.7898} &0.7876    &0.5749 &0.7410 &0.7300    &0.4717&0.6276&0.6410    &0.4402 &0.5977 &0.6113    &0.1543 &0.2856	&0.2673 &352.353\\
        Med-SA \cite{wu2023medical}        &0.5970&0.7405&0.7477	&0.3954&0.7773&0.5667	&0.3007&0.6247&0.4624	&0.3413&0.5424&0.5089	&0.0009&\textbf{0.6513}	&0.0018 &182.901\\
        MA-SAM \cite{chen2024ma}        &0.3235&0.3848&0.4889	    &0.2127&0.2164&0.3507   &0.0508&0.0895&0.0967   &0.1576&0.1858&0.2724   &0.0401&0.0414&0.0772 &33.129\\
        \shline
        MONCH       &\textbf{0.6679}&\textbf{0.7898}&\textbf{0.8009}	&\textbf{0.6572}&\textbf{0.7841}&\textbf{0.7931}	&0.4984&0.6624&0.6652	&\textbf{0.4677}&\textbf{0.6709}&\textbf{0.6373}	&\textbf{0.1767}&0.5194&\textbf{0.3003} &214.456\\
        \shline
    \end{tabular}}
    \label{tab: sota_celltype}
\end{table*}

\subsection{Experimental Setup}
\textbf{Datasets.}
We adopt two well-known pathological cell segmentation datasets.
\textbf{PanNuke~\cite{gamper2019pannuke}} consists of H\&E-stained organ samples from 19 organs, with annotations for various cell types, including neoplastic, epithelial, inflammatory, connective, and dead cells. The dataset contains 189,744 annotated cells, each labeled with both class and shape information. It comprises 7,094 patches captured at 40x magnification, each with a resolution of 256$\times$256 pixels. 
As shown in Fig.~\ref{fig:datadistribution}, PanNuke is highly imbalanced in terms of both cell types and organ types, making it one of the most challenging datasets for cell semantic segmentation.

\noindent\textbf{Implementation details.} The proposed MONCH framework leverages pre-trained text and image encoders from CLIP, with the image encoder further fine-tuned for our specific task. For PanNuke, we use the three-fold splits provided by the dataset organizers to divide the data into training, validation, and testing sets.
We train MONCH for 40 epochs using the Adam optimizer, with an initial learning rate of 5e-5 and a learning rate decay factor of 0.1. 
The text input length for MONCH is set to 77, with an embedding dimension of 1024. 
We implement our method using PyTorch, and all experiments were conducted on NVIDIA GeForce RTX 3090 GPUs.

\begin{figure}
    \centering
    \includegraphics[width=\linewidth]{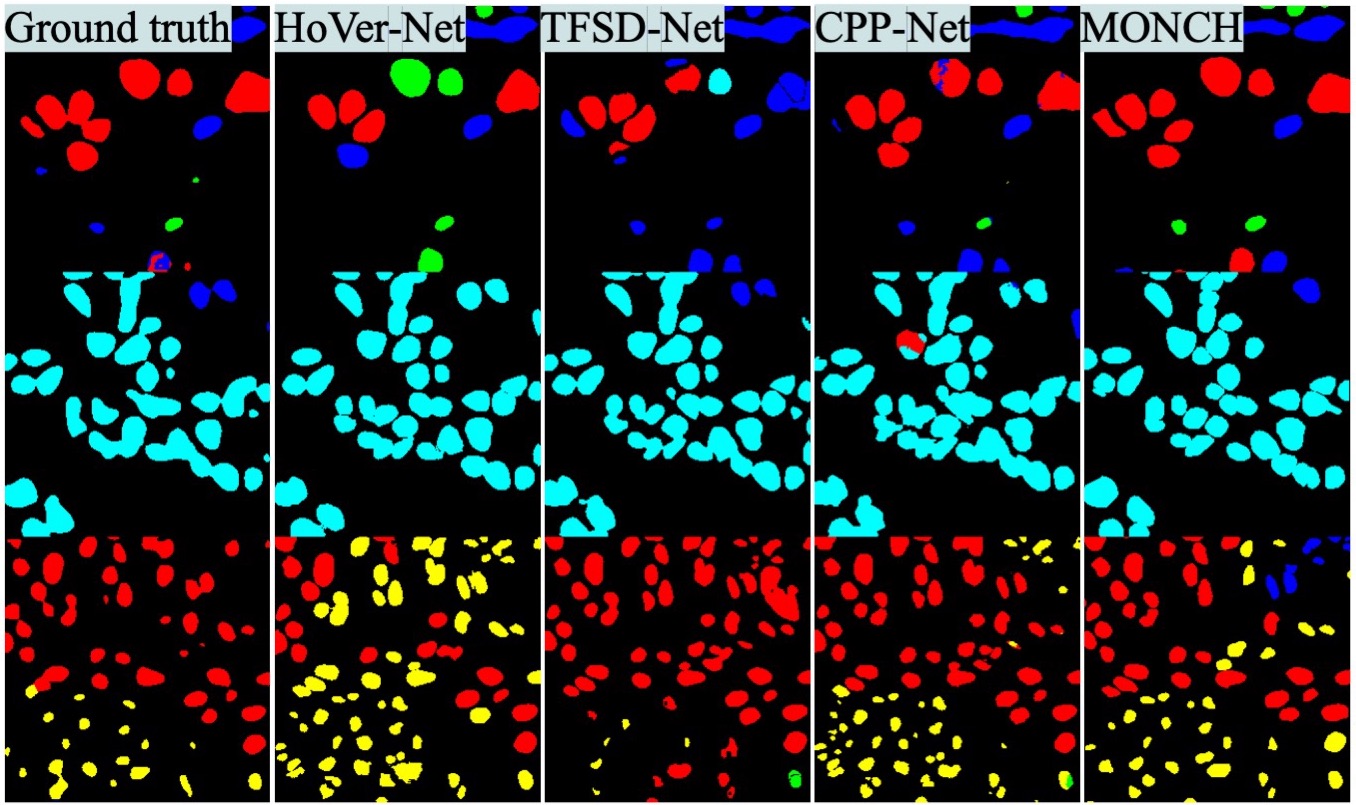}
    \caption{Visualization of multi-organ, multi-cell semantic segmentation in PanNuke.}
    \label{fig:cellseg}
    \vspace{-1mm}
\end{figure}
\noindent\textbf{Metrics.} For evaluation, we report Pixel Accuracy (PA), Intersection over Union (IoU), Frequency Weighted Intersection over Union (FWIoU), Precision, and F1 Score for the multi-class cell semantic segmentation task.

\begin{table*}[t]
    \tablestyle{6pt}{0.9}
        \caption{Evaluation against SOTA cell segmentation methods in organ types from PanNuke. The best results are highlighted in \textbf{bold}.}
        
        \resizebox{\textwidth}{!}{
        \begin{tabular}{*l ^c ^c ^c ^c ^c ^c ^c ^c ^c ^c ^c ^c ^c ^c ^c}
        \shline
        ~   &\multicolumn{3}{c}{Hover-Net \cite{graham2019hover}} &\multicolumn{3}{c}{TSFD-Net \cite{ilyas2022tsfd}} &\multicolumn{3}{c}{CPP-Net \cite{chen2023cpp}} &\multicolumn{3}{c}{MONCH}\\
        \cmidrule(r){2-4} \cmidrule(r){5-7} \cmidrule(r){8-10} \cmidrule(r){11-13}
        ~ &$IoU$ &$FWIOU$ &$F1 Score$ &$IoU$ &$FWIOU$ &$F1 Score$ &$IoU$ &$FWIOU$ &$F1 Score$ &$IoU$ &$FWIOU$ &$F1 Score$ \\
        \shline
        Adrenal     &0.4404 &0.9112 &0.5573    &0.4625 &0.9127 &0.6309     &0.4981&0.9202&0.6543     &0.5104&0.9159&0.6145\\
        Bile Duct   &0.4445 &0.8942 &0.6004    &0.5738 &0.8929 &0.7092     &0.5622&0.8945&0.6973 &0.6164&0.8892&0.7484\\
        Bladder     &0.4377 &0.9242 &0.5618    &0.5166 &0.9190 &0.6156     &0.5154&0.9206&0.6103 &0.5362&0.9169&0.6290\\
        Breast      &0.4898 &0.8747 &0.6046    &0.5254 &0.8808 &0.6907     &0.5154&0.9206&0.6103 &0.5305&0.8793&0.6364\\
        Cervix      &0.4161 &0.8873 &0.5137    &0.4329 &0.8896 &0.5959     &0.4333&0.8812&0.5904 &0.5382&0.8810&0.6421\\
        Colon       &0.5305 &0.8630 &0.6715    &0.5211 &0.8670 &0.6912     &0.5017&0.8661&0.6714 &0.5961&0.8679&0.7434\\
        Esophagus   &0.5090 &0.8830 &0.6349    &0.5069 &0.8816 &0.6725     &0.4733&0.8791&0.6366 &0.5313&0.8713&0.6942\\
        Head \& Neck&0.4630 &0.8961	&0.5915    &0.4844 &0.8989 &0.6516     &0.4780&0.9038&0.6378 &0.5472&0.8913&0.6776\\
        Kidney      &0.3494	&0.9035	&0.4921    &0.4792 &0.9128 &0.6495     &0.5746&0.9187&0.7116 &0.6170&0.9091&0.7480\\
        Liver       &0.4807	&0.9028	&0.6013    &0.5151 &0.9069 &0.6821     &0.4923&0.9071&0.6646 &0.5330&0.9057&0.6381\\
        Lung        &0.4807	&0.9028	&0.6013    &0.3539 &0.8307 &0.5253     &0.3890&0.8367&0.5568 &0.4591&0.8244&0.5941\\
        Ovarian     &0.4941	&0.8412	&0.6754    &0.6208 &0.8440 &0.7563     &0.6507&0.8516&0.7714 &0.6413&0.8352&0.7680\\
        Pancreatic  &0.3910	&0.8641	&0.5907    &0.5191 &0.8681 &0.6806     &0.5315&0.8735&0.6847 &0.6390&0.8804&0.7686\\
        Prostate    &0.4126	&0.8786	&0.5536    &0.5242 &0.8704 &0.6467     &0.4471&0.8657&0.5903 &0.5811&0.8736&0.7204\\
        Skin        &0.3386	&0.8018	&0.4748    &0.4662 &0.8010 &0.6347     &0.4317&0.8147&0.6174 &0.5895&0.8172&0.7202\\
        Stomach     &0.5049	&0.8724	&0.6624    &0.5697 &0.8677 &0.7361     &0.5639&0.8739&0.7292 &0.5461&0.8466&0.6836\\
        Testis      &0.5121	&0.8823	&0.6867    &0.6515 &0.8838 &0.7805     &0.6507&0.8900&0.7792 &0.6622&0.8820&0.7897\\
        Thyroid     &0.4310	&0.8670	&0.5643    &0.4314 &0.8673 &0.6025     &0.4555&0.8778&0.6245 &0.5088&0.8861&0.6754\\
        Uterus      &0.3837	&0.8225	&0.4753    &0.3921 &0.8222 &0.5189     &0.3895&0.8245&0.5183 &0.4610&0.8044&0.6088\\
        \shline
        Average     &0.5203	&0.8738	&0.6563    &0.5282 &\textbf{0.8765} &0.6735     &0.5383&0.8772&0.6680    &\textbf{0.5662}&0.8743&\textbf{0.7035}\\
        STD         &0.0032 &0.0009 &0.0042    &0.0050 &0.0010 &0.0044     &0.0053&0.0009&0.0044	   &0.0032&0.0010&0.0035\\
        \shline
        \end{tabular}
        }
        
    \label{tab: sota_organtype}
\end{table*}

\begin{figure*}[t]
    \centering
    \includegraphics[width=0.97\linewidth]{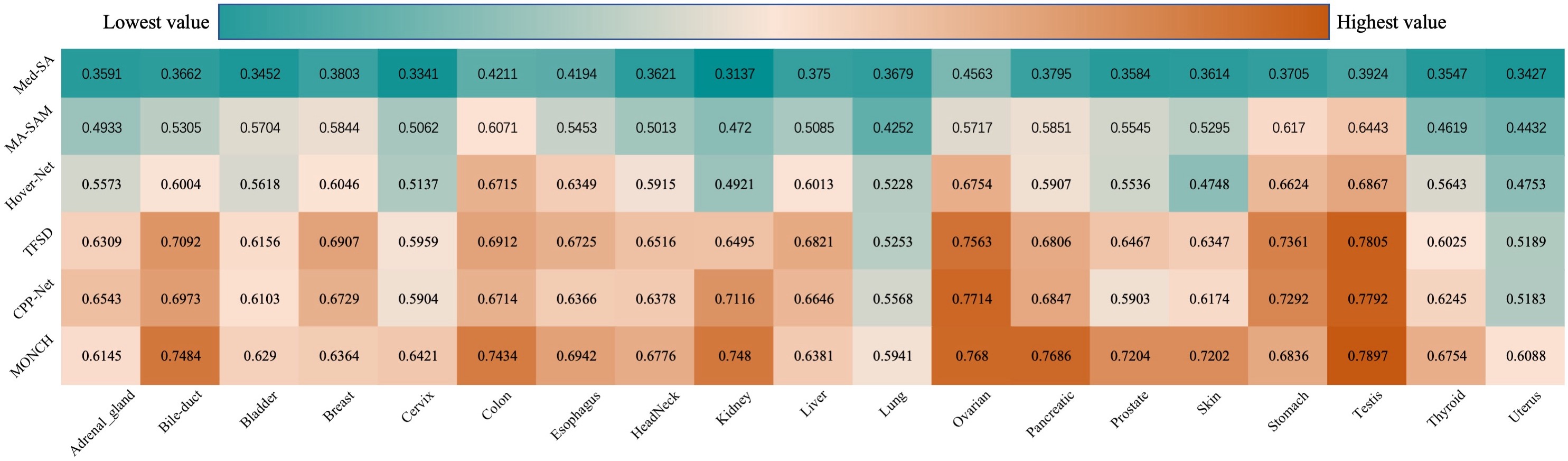}
    \caption{F1 Score of evaluation against SOTA cell segmentation methods in organ types from PanNuke.}
    \vspace{-2mm}
    \label{fig:organ}
\end{figure*}
\subsection{Comparison with State-of-the-Art}
We evaluate MONCH against the state-of-the-art (SOTA) pathological cell segmentation methods that utilize visual inputs, including HoVer-Net \cite{graham2019hover}, TSFD-Net \cite{ilyas2022tsfd}, and CPP-Net \cite{chen2023cpp}. 
As shown in Table~\ref{tab: sota_celltype}, MONCH achieves superior performance across almost all evaluation metrics while maintaining a simpler single-branch architecture.
Notably, our method demonstrates robust performance on PanNuke, as shown in Table~\ref{tab: sota_celltype}, which exhibits significant class imbalance, particularly for epithelial and dead cell categories.
The vision-language approach of MONCH proves especially effective in handling limited data scenarios compared to purely vision-based alternatives.
A key advantage of MONCH lies in its architectural efficiency. While competing methods often employ multi-branch networks, our approach achieves comparable or superior results with a single branch. Its inference time is also comparable with those methods, as shown in Table~\ref{tab: sota_celltype}.
Given that MONCH leverages vision-language model fine-tuning, we also benchmark against SOTA fine-tuned medical image segmentation approaches, specifically Med-SA \cite{wu2023medical} and MA-SAM \cite{chen2024ma} in Table~\ref{tab: sota_celltype}. This comparison provides insights into the effectiveness of our fine-tuning strategy within the medical imaging domain. MA-SAM is specifically designed for medical imaging through fine-tuning SAM, failing to represent pathological data with higher complexity. Med-SA performs much better due to its ability to represent multimodal medical images. Our method combines both textual and multi-grained visual information, resulting in its good characterization capabilities in diverse pathological cell samples. Fig.~\ref{fig:cellseg} illustrates that MONCH can segment the diverse cell dataset much better.

\vspace{-1mm}
The multi-organ composition of PanNuke makes assessing organ-specific performance crucial for validating model robustness.
Table~\ref{tab: sota_organtype} presents the organ-wise comparison between MONCH and SOTA methods.
The results show that MONCH maintains consistent performance across diverse organ types while achieving comparable or superior metrics to existing approaches, indicating its strong generalization capability. Fig.~\ref{fig:organ} intuitively shows the F1 Score of the proposed MONCH against SOTA cell segmentation methods, demonstrating that MONCH can get the best-balanced evaluation results in all organ types.

\begin{table}[t]
    \tablestyle{6pt}{1}
    \caption{F1 Score value comparison of the proposed MONCH with different strategies. MGFE is the multi-grained visual feature extraction block, and PPD is the progressive prompt decoder.}
    \resizebox{\linewidth}{!}{
    \begin{tabular}{^c ^c ^c ^c ^c ^c ^c ^c}
        \shline
        Dataset &MGFE &PPD &Neoplastic &Epithelial &Inflammatory &Connective &Dead\\
        \shline
        \multirow{3}{*}{PanNkue} &$\times$ &\checkmark &0.7720 &0.7566 &0.6125 &0.5959 &0.1333\\
        ~&\checkmark &$\times$ &0.7289 &0.7192 &0.5638 &0.5343 &0.0815\\
        ~&\checkmark &\checkmark &\textbf{0.8009}    &\textbf{0.7931} &\textbf{0.6652}	&\textbf{0.6373}	&\textbf{0.3003}\\
        \shline
    \end{tabular}
    }
    \vspace{-0.5mm}
    \label{tab: strategy}
\end{table}

\subsection{Ablation Studies}
\noindent\textbf{Proposed strategies.} As listed in Table~\ref{tab: strategy}, the proposed MONCH shows the optimal performance when both strategies are implemented concurrently. Without the multi-grained visual feature extraction block, the multi-grained features directly generated from the pre-trained VLM are input to the following progressive prompt decoder. From Table~\ref{tab: strategy}, we can tell that F1 Score values decrease by over 6\% in almost all cell types except epithelial cell because it has adequate data in PanNuke. 
To evaluate the effectiveness of the progressive prompt decoder block, we remove this block out of the proposed method and simply fuse the former multi-grained visual features with the fusion module in Fig.~\ref{fig: framework}. F1 Score values drop by over 9\% in all cell types of PanNuke, proving that iteratively learning from the fine feature to coarse feature, finally merged with a blended feature, can effectively integrate multimodal features. 
It is obvious in Table~\ref{tab: strategy} that training without each one of the modules will lead to quite large performance degradation in dead cell, proving that both modules can enhance the model robustness in imbalanced dataset.

\begin{table}[t]
    \tablestyle{6pt}{1}
    \caption{F1 Score value comparison of the proposed MONCH with different visual feature extraction strategies. HF represents a high-frequency feature enhancement module. Conv represents the convolution module, and Topo represents the topological structure enhancement module. $\times$ symbol means that the feature extraction block is replaced with the convolution block.}
    \resizebox{\linewidth}{!}{
    \begin{tabular}{^c ^c ^c ^c ^c ^c ^c ^c ^c}
        \shline
        Dataset &HF &Conv &Topo &Neoplastic &Epithelial &Inflammatory &Connective &Dead\\
        \shline
        \multirow{3}{*}{PanNuke}
        &$\times$ &\checkmark &\checkmark &0.8004 &0.7888 &0.6630 &0.6380 &0.2449\\
        ~&\checkmark &\checkmark &$\times$ &0.7915 &0.7806 &0.6498 &0.6322 &0.3041\\
        ~&\checkmark &\checkmark &\checkmark &\textbf{0.8009}    &\textbf{0.7931} &\textbf{0.6652}	&\textbf{0.6373}	&\textbf{0.3003}\\
        \shline
    \end{tabular}}
    \label{tab: visual}
\end{table}

\vspace{1mm}
\noindent\textbf{Visual feature extraction setting.} To evaluate the necessity of multi-grained visual feature extraction modules, we conduct ablation experiments and F1 Score values are listed in Table~\ref{tab: visual}. Compared with simple convolution feature, adding any complementary visual feature can achieve performance improvements. From this table, we can find that both high-frequency feature extraction block and topological structure enhancement block are important for dead cell image feature representation with very limited data.

\begin{table}[t]
    \tablestyle{6pt}{1.1}
    \caption{F1 Score value comparison of the proposed MONCH with different prompt decoders. $\{q_h, q_v, q_t, q_T\}$ represent the queries for the next-level feature to conduct multi-head self-attention in Fig.~\ref{fig: framework}. $\times$ symbol in this table means that the feature from the previous level are carried over to the multi-head-self-attention mechanism at the next level.}
    \resizebox{\linewidth}{!}{
    \begin{tabular}{^c ^c ^c ^c  ^c ^c ^c ^c ^c}
        \shline
        $q_h$ &$q_v$ &$q_t$ &$q_T$ &Neoplastic &Epithelial &Inflammatory &Connective &Dead\\
        \shline
        $\times$ &\checkmark &\checkmark &\checkmark &0.7890 &0.7806 &0.6460 &0.6213 &0.2136\\
        \checkmark &$\times$ &\checkmark &\checkmark &0.7912 &0.7800 &0.6477 &0.6221 &0.3001\\
        \checkmark &\checkmark &$\times$ &\checkmark  &0.7895 &0.7744 &0.6489 &0.6062 &0.1804\\
        \checkmark &\checkmark &\checkmark &$\times$  &0.7750 &0.7643 &0.6273 &0.5781 &0.0620\\
        \checkmark &\checkmark &\checkmark &\checkmark &\textbf{0.8009}    &\textbf{0.7931} &\textbf{0.6652}	&\textbf{0.6373}	&\textbf{0.3003}\\
        \shline
    \end{tabular}
    }
    \label{tab: PPD}
\end{table}

\noindent\textbf{Progressive prompt decoder setting.} We evaluate our proposed method with different progressive prompt decoders, i.e., separately removing each one of the stages in $\{q_h, q_v, q_t, q_T\}$. Table~\ref{tab: PPD} demonstrates that removing any one of these stages will significantly weaken the model's performance. The deeper the stage is, the greater the degradation, which illustrates that the finer feature can indeed provide reliable information for the coarser feature. 

\begin{table}[t]
    \tablestyle{6pt}{1.1}
    \caption{F1 Score value comparison of the proposed MONCH with different pre-trained backbones on PanNuke.}
    \resizebox{\linewidth}{!}{
    \begin{tabular}{^c ^c  ^c ^c ^c ^c ^c}
        \shline
        Backbone &Neoplastic &Epithelial &Inflammatory &Connective &Dead\\
        \shline
        PLIP &0.7346	&0.7212	&0.4864	&0.5161	&0.1364\\
        CONCH &0.7892	&0.7779	&0.6382	&0.6102	&0.3237\\
        CLIP &\textbf{0.8009}    &\textbf{0.7931} &\textbf{0.6652}	&\textbf{0.6373}	&\textbf{0.3003}\\
        \shline
    \end{tabular}
    }
    \label{tab: backbones}
\end{table}
\vspace{1mm}
\noindent\textbf{Different backbones.} 
We evaluate MONCH using three different vision-language models (VLMs) as backbones: CLIP, PLIP, and CONCH, on PanNuke. Table~\ref{tab: backbones} shows the F1 scores for each cell type, averaged across all organ types. CLIP-based MONCH achieves the best overall performance, with CONCH-based implementation showing competitive results. However, PLIP-based MONCH shows notably lower performance, particularly for cell types with limited training samples (Inflammatory, Connective, and Dead cells). We attribute this performance gap to CONCH's architectural design, which was originally optimized for whole slide image (WSI) classification. Its pre-trained image encoder produces feature maps with approximately half the resolution compared to CLIP's encoder, impacting the model's ability to capture fine-grained cellular details.

We will provide feature maps of different enhancement modules and more visualized results in the supplementary.
\section{Conclusion}
\label{sec:conclusion}
In this paper, we propose a novel progressive multi-modal prompt learning method with a single-branch architecture for multi-organ, multi-class cell semantic segmentation, achieving coarse-to-fine-grained feature extraction using text-image pairs.
Specifically, our single-branch network effectively analyzes multimodal pathological cell features.
The multi-grained visual feature extraction module enhances visual features from coarse to fine level.
Subsequently, the progressive prompt decoder fully integrates these multimodal features through a sequence of fine-coarse-fine queries, enabling the multi-head self-attention modules to refine and improve feature representations at the next level.
Evaluations conducted on a complex cell segmentation dataset demonstrate that our proposed method outperforms state-of-the-art cell segmentation techniques and vision-language models in semantic segmentation tasks, highlighting its effectiveness in capturing intricate cellular structures.
\clearpage
\setcounter{page}{1}

\twocolumn[{
    \renewcommand\twocolumn[1][]{#1}
    \begin{center}
        \centering
        \maketitlesupplementary
        \includegraphics[width=\linewidth]{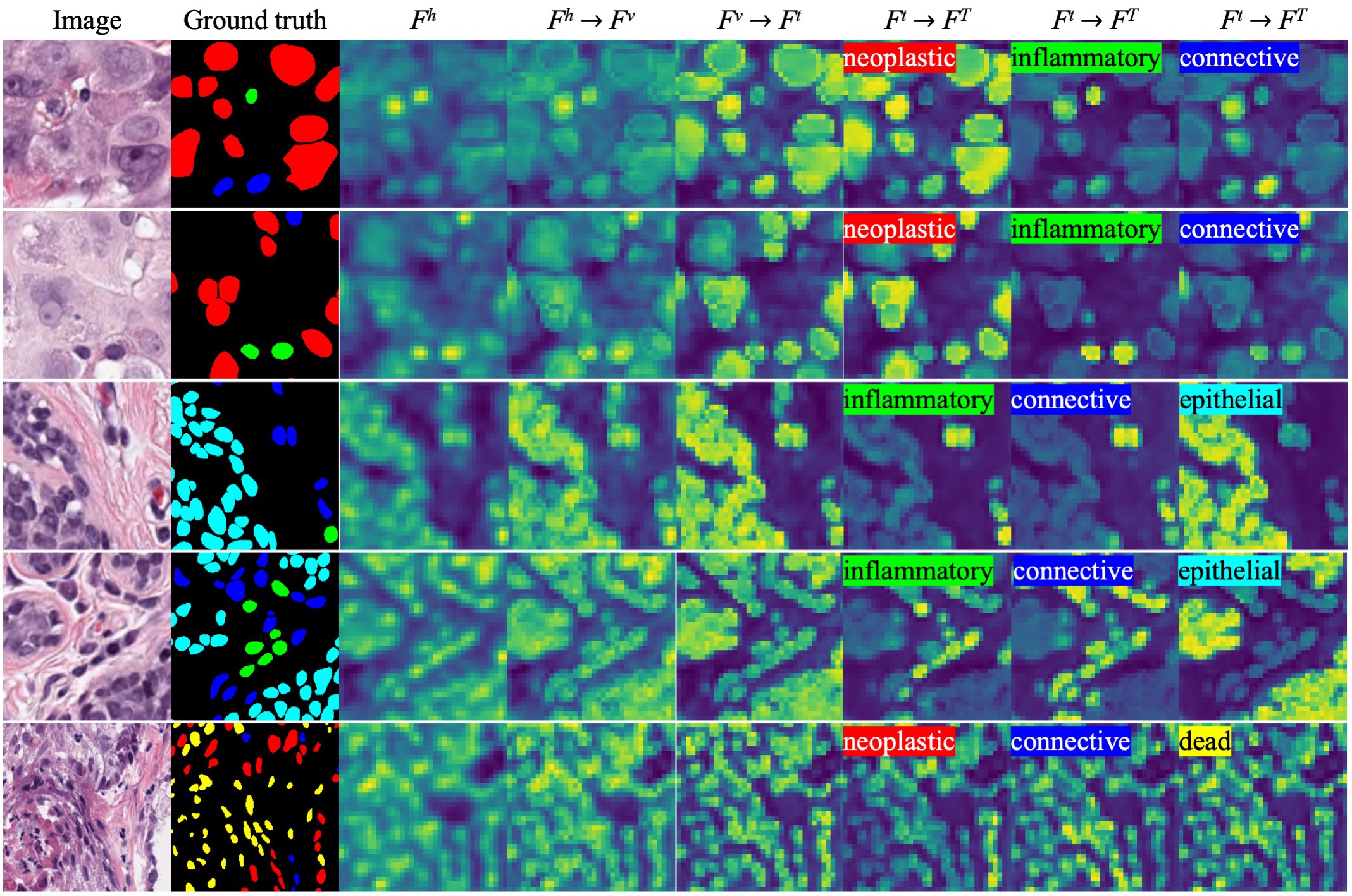}
        \vspace{-0.9cm}
         \captionof{figure}{Visualization of multi-grained visual features and multimodal features in progressive prompt decoder block. Multimodal features can emphasize cells of specific types as guided by linguistic prompts.}
        \label{fig:feature}
    \end{center}
}]

\section{Additionally Results}
\label{sec:rationale}

\subsection{Different backbones} 
We evaluate MONCH with three vision-language models (VLMs) as backbones: PLIP \cite{huang2023visual}, CONCH \cite{lu2024visual}, and CLIP \cite{radford2021learning}. 
Additionally, we replace the image encoder with three vision-based models, SAM \cite{kirillov2023segment} and UNI \cite{chen2024towards}, as well as SAM2 \cite{ravi2024sam}, paired with a text encoder from pre-trained CLIP.
Table~\ref{tab: allbackbones} reports the F1 scores for each cell type, averaged across all organ types.

Among these, CLIP-based MONCH achieves the highest overall performance, followed by competitive results from SAM-based, CONCH-based, and SAM2-based implementations.
PLIP-based MONCH shows a significantly lower performance, particularly for cell types with limited training samples.
This gap can be attributed to PLIP’s architectural design, which was optimized for whole-slide image (WSI) classification.
PLIP’s pre-trained image encoder generates feature maps with approximately half the resolution compared to the CLIP encoder, reducing the model's ability to capture fine-grained cellular details. 
Furthermore, MONCH with UNI-based backbones demonstrates lower performance, likely due to the misalignment between text and image encoders.

\begin{table}[t]
    \tablestyle{6pt}{1.1}
    \caption{F1 Score value comparison of the proposed MONCH with different pre-trained backbones on PanNuke.}
    \resizebox{\linewidth}{!}{
    \begin{tabular}{^c ^c ^c  ^c ^c ^c ^c ^c}
        \shline
        Backbone &Pretraining &Neoplastic &Epithelial &Inflammatory &Connective &Dead\\
        \shline
        SAM \cite{kirillov2023segment} &Vision &0.7952 &0.7857 &0.6542 &0.6271 &0.2038\\
        UNI \cite{chen2024towards} &Vision &0.7451	&0.7452	&0.5837	&0.5606	&0.2309\\
        SAM2 \cite{ravi2024sam} &Vision &0.7960	&0.7856	&0.6553	&0.6311	&0.2918\\
        PLIP \cite{huang2023visual} &Vision-language &0.7346	&0.7212	&0.4864	&0.5161	&0.1364\\
        CONCH \cite{lu2024visual}  &Vision-language &0.7892	&0.7779	&0.6382	&0.6102	&0.3237\\
        CLIP \cite{radford2021learning} &Vision-language &\textbf{0.8009}    &\textbf{0.7931} &\textbf{0.6652}	&\textbf{0.6373}	&\textbf{0.3003}\\
        \shline
    \end{tabular}
    }
    \label{tab: allbackbones}
\end{table}

\begin{figure*}[t]
    \centering
    \includegraphics[width=\linewidth]{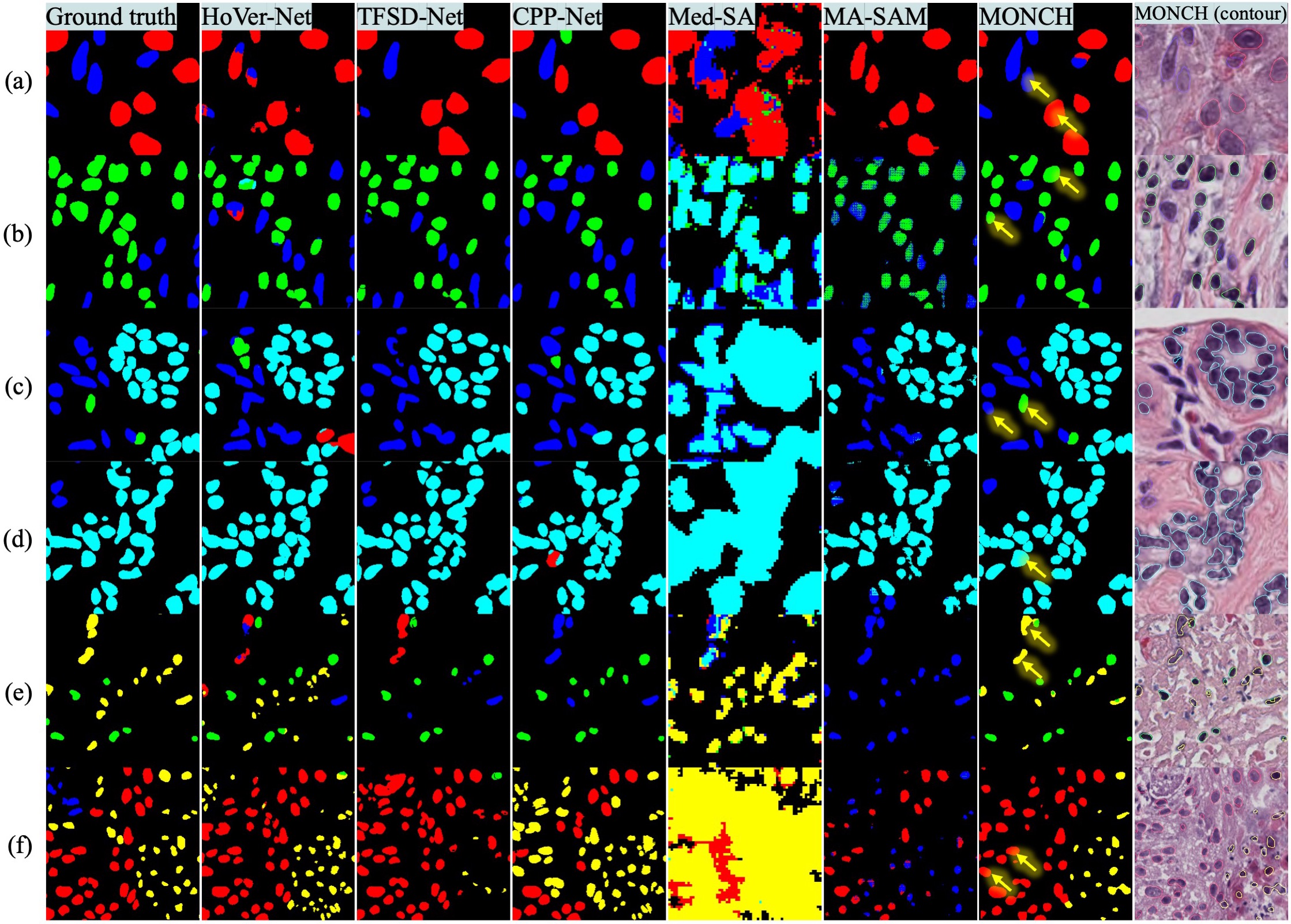}
    \vspace{-0.7cm}
    \caption{Visualization of multi-organ, multi-cell semantic segmentation in PanNuke. \textcolor{red}{Red} represents neoplastic cell, \textcolor{green}{green} represents inflammatory cell, \textcolor{blue}{blue} represents connective cell, \textcolor{cyan}{cyan} represents epithelial cell, and \textcolor{yellow}{yellow} represents dead cell. Cells are outlined with contours in their respective annotation colors of MONCH.}
    \label{fig:cellsegnew}
    \vspace{-0.3cm}
\end{figure*}

\subsection{Feature visualization} 
The progressive prompt decoder block plays a crucial role in harmonizing linguistic and visual features.
To demonstrate its effectiveness, we visualize the iteratively generated visual and multimodal features in Fig.~\ref{fig:feature}.
%
The high-frequency visual features, $\mathbf{F}^h$, effectively capture detailed information about pathological cells.
By utilizing these fine-grained features as the $query$ for subsequent coarse features, the visual representations progressively refine, as shown by $\mathbf{F}^h \to \mathbf{F}^v \to \mathbf{F}^t$.
These multi-grained visual features excel at preserving the semantic information of pathological cell textual features.
Subsequently, the visual features are used as $query$ inputs to the text features, facilitating the capture of multimodal features that integrate linguistic and visual data.
%
Finally, by merging visual and textual features, the multimodal features, $\mathbf{F}^t \to \mathbf{F}^\mathcal{T}$, effectively highlight features specific to different cell types.

\subsection{Visualization of cell segmentation} 
Fig.~\ref{fig:cellsegnew} presents the cell segmentation results of MONCH compared to state-of-the-art cell segmentation methods and large-scale models.
%
MONCH clearly outperforms competing approaches,  leveraging both linguistic and visual features to effectively segment diverse cell data.
%
Notably, MONCH delivers superior semantic segmentation performance compared to CPP-Net, the second-best performing method. 
By integrating linguistic information with visual learning, MONCH accurately identifies cell types, overcoming challenges faced by other methods.
%
As shown in Fig.~\ref{fig:cellsegnew}, cells marked with yellow arrows are misclassified by CPP-Net but are correctly segmented by MONCH.
%
Furthermore, MONCH effectively segments rare instances, such as dead cells in PanNuke, despite limited data available. 
This underscores MONCH's robustness in handling imbalanced datasets, making it a powerful solution for challenging segmentation tasks.
%
%
\clearpage
{
    \small
    \bibliographystyle{ieeenat_fullname}
    \bibliography{main}

\begin{thebibliography}{54}
\providecommand{\natexlab}[1]{#1}
\providecommand{\url}[1]{\texttt{#1}}
\expandafter\ifx\csname urlstyle\endcsname\relax
  \providecommand{\doi}[1]{doi: #1}\else
  \providecommand{\doi}{doi: \begingroup \urlstyle{rm}\Url}\fi

\bibitem[Abousamra et~al.(2021)Abousamra, Belinsky, Van~Arnam, Allard, Yee, Gupta, Kurc, Samaras, Saltz, and Chen]{abousamra2021multi}
Shahira Abousamra, David Belinsky, John Van~Arnam, Felicia Allard, Eric Yee, Rajarsi Gupta, Tahsin Kurc, Dimitris Samaras, Joel Saltz, and Chao Chen.
\newblock Multi-class cell detection using spatial context representation.
\newblock In \emph{ICCV}, pages 4005--4014, 2021.

\bibitem[Berger et~al.(2024)Berger, Lux, Stucki, B{\"u}rgin, Shit, Banaszak, Rueckert, Bauer, and Paetzold]{berger2024topologically}
Alexander~H Berger, Laurin Lux, Nico Stucki, Vincent B{\"u}rgin, Suprosanna Shit, Anna Banaszak, Daniel Rueckert, Ulrich Bauer, and Johannes~C Paetzold.
\newblock Topologically faithful multi-class segmentation in medical images.
\newblock In \emph{MICCAI}, pages 721--731. Springer, 2024.

\bibitem[Bokhorst et~al.(2023)Bokhorst, Nagtegaal, Fraggetta, Vatrano, Mesker, Vieth, van~der Laak, and Ciompi]{bokhorst2023deep}
John-Melle Bokhorst, Iris~D Nagtegaal, Filippo Fraggetta, Simona Vatrano, Wilma Mesker, Michael Vieth, Jeroen van~der Laak, and Francesco Ciompi.
\newblock Deep learning for multi-class semantic segmentation enables colorectal cancer detection and classification in digital pathology images.
\newblock \emph{Scientific Reports}, 13\penalty0 (1):\penalty0 8398, 2023.

\bibitem[Chen et~al.(2024{\natexlab{a}})Chen, Miao, Wu, Zhong, Yan, Kim, Hu, Liu, Sun, Li, et~al.]{chen2024ma}
Cheng Chen, Juzheng Miao, Dufan Wu, Aoxiao Zhong, Zhiling Yan, Sekeun Kim, Jiang Hu, Zhengliang Liu, Lichao Sun, Xiang Li, et~al.
\newblock Ma-sam: Modality-agnostic sam adaptation for 3d medical image segmentation.
\newblock \emph{Medical Image Analysis}, 98:\penalty0 103310, 2024{\natexlab{a}}.

\bibitem[Chen et~al.(2021)Chen, Fan, and Panda]{chen2021crossvit}
Chun-Fu~Richard Chen, Quanfu Fan, and Rameswar Panda.
\newblock Crossvit: Cross-attention multi-scale vision transformer for image classification.
\newblock In \emph{CVPR}, pages 357--366, 2021.

\bibitem[Chen et~al.(2024{\natexlab{b}})Chen, Ding, Lu, Williamson, Jaume, Song, Chen, Zhang, Shao, Shaban, et~al.]{chen2024towards}
Richard~J Chen, Tong Ding, Ming~Y Lu, Drew~FK Williamson, Guillaume Jaume, Andrew~H Song, Bowen Chen, Andrew Zhang, Daniel Shao, Muhammad Shaban, et~al.
\newblock Towards a general-purpose foundation model for computational pathology.
\newblock \emph{Nature Medicine}, 30\penalty0 (3):\penalty0 850--862, 2024{\natexlab{b}}.

\bibitem[Chen et~al.(2023)Chen, Ding, Liu, Cheng, and Tao]{chen2023cpp}
Shengcong Chen, Changxing Ding, Minfeng Liu, Jun Cheng, and Dacheng Tao.
\newblock Cpp-net: Context-aware polygon proposal network for nucleus segmentation.
\newblock \emph{IEEE Transactions on Image Processing}, 32:\penalty0 980--994, 2023.

\bibitem[Dai et~al.(2021)Dai, Gieseke, Oehmcke, Wu, and Barnard]{dai2021attentional}
Yimian Dai, Fabian Gieseke, Stefan Oehmcke, Yiquan Wu, and Kobus Barnard.
\newblock Attentional feature fusion.
\newblock In \emph{WACV}, pages 3560--3569, 2021.

\bibitem[Dogar et~al.(2023)Dogar, Shahzad, and Fraz]{dogar2023attention}
G~Murtaza Dogar, Muhammad Shahzad, and Muhammad~Moazam Fraz.
\newblock Attention augmented distance regression and classification network for nuclei instance segmentation and type classification in histology images.
\newblock \emph{Biomedical Signal Processing and Control}, 79:\penalty0 104199, 2023.

\bibitem[Fehri et~al.(2019)Fehri, Gooya, Lu, Meijering, Johnston, and Frangi]{fehri2019bayesian}
Hamid Fehri, Ali Gooya, Yuanjun Lu, Erik Meijering, Simon~A Johnston, and Alejandro~F Frangi.
\newblock Bayesian polytrees with learned deep features for multi-class cell segmentation.
\newblock \emph{IEEE Transactions on Image Processing}, 28\penalty0 (7):\penalty0 3246--3260, 2019.

\bibitem[Gamper et~al.(2019)Gamper, Alemi~Koohbanani, Benet, Khuram, and Rajpoot]{gamper2019pannuke}
Jevgenij Gamper, Navid Alemi~Koohbanani, Ksenija Benet, Ali Khuram, and Nasir Rajpoot.
\newblock Pannuke: an open pan-cancer histology dataset for nuclei instance segmentation and classification.
\newblock In \emph{ECDP 2019}, pages 11--19. Springer, 2019.

\bibitem[Gamper et~al.(2020)Gamper, Koohbanani, Benes, Graham, Jahanifar, Khurram, Azam, Hewitt, and Rajpoot]{gamper2020pannuke}
Jevgenij Gamper, Navid~Alemi Koohbanani, Ksenija Benes, Simon Graham, Mostafa Jahanifar, Syed~Ali Khurram, Ayesha Azam, Katherine Hewitt, and Nasir Rajpoot.
\newblock Pannuke dataset extension, insights and baselines.
\newblock \emph{arXiv preprint arXiv:2003.10778}, 2020.

\bibitem[Gao et~al.(2024)Gao, Shi, Zhu, Belval, Nuriel, Appalaraju, Ghadar, Tu, Mahadevan, and Soatto]{gao2024enhancing}
Yuan Gao, Kunyu Shi, Pengkai Zhu, Edouard Belval, Oren Nuriel, Srikar Appalaraju, Shabnam Ghadar, Zhuowen Tu, Vijay Mahadevan, and Stefano Soatto.
\newblock Enhancing vision-language pre-training with rich supervisions.
\newblock In \emph{CVPR}, pages 13480--13491, 2024.

\bibitem[Graham et~al.(2019)Graham, Vu, Raza, Azam, Tsang, Kwak, and Rajpoot]{graham2019hover}
Simon Graham, Quoc~Dang Vu, Shan E~Ahmed Raza, Ayesha Azam, Yee~Wah Tsang, Jin~Tae Kwak, and Nasir Rajpoot.
\newblock Hover-net: Simultaneous segmentation and classification of nuclei in multi-tissue histology images.
\newblock \emph{Medical Image Analysis}, 58:\penalty0 101563, 2019.

\bibitem[Guo et~al.(2020)Guo, Fan, Zhang, Xiang, and Pan]{guo2020augfpn}
Chaoxu Guo, Bin Fan, Qian Zhang, Shiming Xiang, and Chunhong Pan.
\newblock Augfpn: Improving multi-scale feature learning for object detection.
\newblock In \emph{CVPR}, pages 12595--12604, 2020.

\bibitem[Han et~al.(2022)Han, Yao, Zhao, Li, Shi, Wu, Chen, Qu, Zhao, Lan, et~al.]{han2022meta}
Chu Han, Huasheng Yao, Bingchao Zhao, Zhenhui Li, Zhenwei Shi, Lei Wu, Xin Chen, Jinrong Qu, Ke Zhao, Rushi Lan, et~al.
\newblock Meta multi-task nuclei segmentation with fewer training samples.
\newblock \emph{Medical Image Analysis}, 80:\penalty0 102481, 2022.

\bibitem[Hao et~al.(2024)Hao, Gong, Zeng, Liu, Guo, Cheng, Wang, Ma, Zhang, and Song]{hao2024large}
Minsheng Hao, Jing Gong, Xin Zeng, Chiming Liu, Yucheng Guo, Xingyi Cheng, Taifeng Wang, Jianzhu Ma, Xuegong Zhang, and Le Song.
\newblock Large-scale foundation model on single-cell transcriptomics.
\newblock \emph{Nature Methods}, pages 1--11, 2024.

\bibitem[He et~al.(2023{\natexlab{a}})He, Wang, Wei, Xu, Ji, Liu, and Chen]{he2023toposeg}
Hongliang He, Jun Wang, Pengxu Wei, Fan Xu, Xiangyang Ji, Chang Liu, and Jie Chen.
\newblock Toposeg: Topology-aware nuclear instance segmentation.
\newblock In \emph{CVPR}, pages 21307--21316, 2023{\natexlab{a}}.

\bibitem[He et~al.(2023{\natexlab{b}})He, Unberath, Ke, and Shen]{he2023transnuseg}
Zhenqi He, Mathias Unberath, Jing Ke, and Yiqing Shen.
\newblock Transnuseg: A lightweight multi-task transformer for nuclei segmentation.
\newblock In \emph{MICCAI}, pages 206--215. Springer, 2023{\natexlab{b}}.

\bibitem[Hou et~al.(2021)Hou, Qin, Xiang, Tan, and Xiong]{hou2021af}
Guimin Hou, Jiaohua Qin, Xuyu Xiang, Yun Tan, and Neal~N Xiong.
\newblock Af-net: A medical image segmentation network based on attention mechanism and feature fusion.
\newblock \emph{Computers, Materials \& Continua}, 69\penalty0 (2):\penalty0 1877--1891, 2021.

\bibitem[Huang et~al.(2023)Huang, Bianchi, Yuksekgonul, Montine, and Zou]{huang2023visual}
Zhi Huang, Federico Bianchi, Mert Yuksekgonul, Thomas~J Montine, and James Zou.
\newblock A visual--language foundation model for pathology image analysis using medical twitter.
\newblock \emph{Nature Medicine}, 29\penalty0 (9):\penalty0 2307--2316, 2023.

\bibitem[Huo et~al.(2024)Huo, Sun, Tian, Wang, Yu, Long, Zhang, and Li]{huo2024hifuse}
Xiangzuo Huo, Gang Sun, Shengwei Tian, Yan Wang, Long Yu, Jun Long, Wendong Zhang, and Aolun Li.
\newblock Hifuse: Hierarchical multi-scale feature fusion network for medical image classification.
\newblock \emph{Biomedical Signal Processing and Control}, 87:\penalty0 105534, 2024.

\bibitem[Ilyas et~al.(2022)Ilyas, Mannan, Khan, Azam, Kim, and De~Boer]{ilyas2022tsfd}
Talha Ilyas, Zubaer~Ibna Mannan, Abbas Khan, Sami Azam, Hyongsuk Kim, and Friso De~Boer.
\newblock Tsfd-net: Tissue specific feature distillation network for nuclei segmentation and classification.
\newblock \emph{Neural Networks}, 151:\penalty0 1--15, 2022.

\bibitem[Ju et~al.(2021)Ju, Luo, Wang, and Luo]{ju2021adaptive}
Moran Ju, Jiangning Luo, Zhongbo Wang, and Haibo Luo.
\newblock Adaptive feature fusion with attention mechanism for multi-scale target detection.
\newblock \emph{Neural Computing and Applications}, 33:\penalty0 2769--2781, 2021.

\bibitem[Kassim et~al.(2020)Kassim, Palaniappan, Yang, Poostchi, Palaniappan, Maude, Antani, and Jaeger]{kassim2020clustering}
Yasmin~M Kassim, Kannappan Palaniappan, Feng Yang, Mahdieh Poostchi, Nila Palaniappan, Richard~J Maude, Sameer Antani, and Stefan Jaeger.
\newblock Clustering-based dual deep learning architecture for detecting red blood cells in malaria diagnostic smears.
\newblock \emph{IEEE Journal of Biomedical and Health Informatics}, 25\penalty0 (5):\penalty0 1735--1746, 2020.

\bibitem[Kirillov et~al.(2023)Kirillov, Mintun, Ravi, Mao, Rolland, Gustafson, Xiao, Whitehead, Berg, Lo, et~al.]{kirillov2023segment}
Alexander Kirillov, Eric Mintun, Nikhila Ravi, Hanzi Mao, Chloe Rolland, Laura Gustafson, Tete Xiao, Spencer Whitehead, Alexander~C Berg, Wan-Yen Lo, et~al.
\newblock Segment anything.
\newblock In \emph{ICCV}, pages 4015--4026, 2023.

\bibitem[Liang et~al.(2023)Liang, Zhao, Liang, Li, Wu, and Zhou]{liang2023maxformer}
Zhiwei Liang, Kui Zhao, Gang Liang, Siyu Li, Yifei Wu, and Yiping Zhou.
\newblock Maxformer: Enhanced transformer for medical image segmentation with multi-attention and multi-scale features fusion.
\newblock \emph{Knowledge-Based Systems}, 280:\penalty0 110987, 2023.

\bibitem[Liu et~al.(2021)Liu, Fan, Jiang, Liu, and Luo]{liu2021learning}
Jinyuan Liu, Xin Fan, Ji Jiang, Risheng Liu, and Zhongxuan Luo.
\newblock Learning a deep multi-scale feature ensemble and an edge-attention guidance for image fusion.
\newblock \emph{IEEE Transactions on Circuits and Systems for Video Technology}, 32\penalty0 (1):\penalty0 105--119, 2021.

\bibitem[Lu et~al.(2024)Lu, Chen, Williamson, Chen, Liang, Ding, Jaume, Odintsov, Le, Gerber, et~al.]{lu2024visual}
Ming~Y Lu, Bowen Chen, Drew~FK Williamson, Richard~J Chen, Ivy Liang, Tong Ding, Guillaume Jaume, Igor Odintsov, Long~Phi Le, Georg Gerber, et~al.
\newblock A visual-language foundation model for computational pathology.
\newblock \emph{Nature Medicine}, 30\penalty0 (3):\penalty0 863--874, 2024.

\bibitem[Meng et~al.(2018)Meng, Lam, Tsia, and So]{meng2018large}
Nan Meng, Edmund~Y Lam, Kevin~K Tsia, and Hayden Kwok-Hay So.
\newblock Large-scale multi-class image-based cell classification with deep learning.
\newblock \emph{IEEE Journal of Biomedical and Health Informatics}, 23\penalty0 (5):\penalty0 2091--2098, 2018.

\bibitem[Minaee et~al.(2021)Minaee, Boykov, Porikli, Plaza, Kehtarnavaz, and Terzopoulos]{minaee2021image}
Shervin Minaee, Yuri Boykov, Fatih Porikli, Antonio Plaza, Nasser Kehtarnavaz, and Demetri Terzopoulos.
\newblock Image segmentation using deep learning: A survey.
\newblock \emph{IEEE Transactions on Pattern Analysis and Machine Intelligence}, 44\penalty0 (7):\penalty0 3523--3542, 2021.

\bibitem[Oh and Jeong(2024)]{oh2024controllable}
Hyun-Jic Oh and Won-Ki Jeong.
\newblock Controllable and efficient multi-class pathology nuclei data augmentation uuing text-conditioned diffusion models.
\newblock In \emph{MICCAI}, pages 36--46. Springer, 2024.

\bibitem[Pan et~al.(2023)Pan, Cheng, Hou, Lan, Lu, Li, Feng, Wang, Liang, Liu, et~al.]{pan2023smile}
Xipeng Pan, Jijun Cheng, Feihu Hou, Rushi Lan, Cheng Lu, Lingqiao Li, Zhengyun Feng, Huadeng Wang, Changhong Liang, Zhenbing Liu, et~al.
\newblock Smile: Cost-sensitive multi-task learning for nuclear segmentation and classification with imbalanced annotations.
\newblock \emph{Medical Image Analysis}, 88:\penalty0 102867, 2023.

\bibitem[Petukhov et~al.(2022)Petukhov, Xu, Soldatov, Cadinu, Khodosevich, Moffitt, and Kharchenko]{petukhov2022cell}
Viktor Petukhov, Rosalind~J Xu, Ruslan~A Soldatov, Paolo Cadinu, Konstantin Khodosevich, Jeffrey~R Moffitt, and Peter~V Kharchenko.
\newblock Cell segmentation in imaging-based spatial transcriptomics.
\newblock \emph{Nature Biotechnology}, 40\penalty0 (3):\penalty0 345--354, 2022.

\bibitem[Punn and Agarwal(2020)]{punn2020inception}
Narinder~Singh Punn and Sonali Agarwal.
\newblock Inception u-net architecture for semantic segmentation to identify nuclei in microscopy cell images.
\newblock \emph{ACM Transactions on Multimedia Computing, Communications, and Applications}, 16\penalty0 (1):\penalty0 1--15, 2020.

\bibitem[Radford et~al.(2021)Radford, Kim, Hallacy, Ramesh, Goh, Agarwal, Sastry, Askell, Mishkin, Clark, et~al.]{radford2021learning}
Alec Radford, Jong~Wook Kim, Chris Hallacy, Aditya Ramesh, Gabriel Goh, Sandhini Agarwal, Girish Sastry, Amanda Askell, Pamela Mishkin, Jack Clark, et~al.
\newblock Learning transferable visual models from natural language supervision.
\newblock In \emph{ICML}, pages 8748--8763. PMLR, 2021.

\bibitem[Ravi et~al.(2024)Ravi, Gabeur, Hu, Hu, Ryali, Ma, Khedr, R{\"a}dle, Rolland, Gustafson, et~al.]{ravi2024sam}
Nikhila Ravi, Valentin Gabeur, Yuan-Ting Hu, Ronghang Hu, Chaitanya Ryali, Tengyu Ma, Haitham Khedr, Roman R{\"a}dle, Chloe Rolland, Laura Gustafson, et~al.
\newblock Sam 2: Segment anything in images and videos.
\newblock \emph{arXiv preprint arXiv:2408.00714}, 2024.

\bibitem[Tong et~al.(2023)Tong, Su, Wu, Guo, Wei, Zuo, and Sun]{tong2023msaffnet}
Xiaozhong Tong, Shaojing Su, Peng Wu, Runze Guo, Junyu Wei, Zhen Zuo, and Bei Sun.
\newblock Msaffnet: A multiscale label-supervised attention feature fusion network for infrared small target detection.
\newblock \emph{IEEE Transactions on Geoscience and Remote Sensing}, 61:\penalty0 1--16, 2023.

\bibitem[Verma et~al.(2021)Verma, Kumar, Patil, Kurian, Rane, Graham, Vu, Zwager, Raza, Rajpoot, et~al.]{verma2021monusac2020}
Ruchika Verma, Neeraj Kumar, Abhijeet Patil, Nikhil~Cherian Kurian, Swapnil Rane, Simon Graham, Quoc~Dang Vu, Mieke Zwager, Shan E~Ahmed Raza, Nasir Rajpoot, et~al.
\newblock Monusac2020: A multi-organ nuclei segmentation and classification challenge.
\newblock \emph{IEEE Transactions on Medical Imaging}, 40\penalty0 (12):\penalty0 3413--3423, 2021.

\bibitem[Wang et~al.(2022)Wang, Cao, Wang, and Zaiane]{wang2022uctransnet}
Haonan Wang, Peng Cao, Jiaqi Wang, and Osmar~R Zaiane.
\newblock Uctransnet: rethinking the skip connections in u-net from a channel-wise perspective with transformer.
\newblock In \emph{AAAI}, pages 2441--2449, 2022.

\bibitem[Wang et~al.(2024)Wang, Wang, Peng, and Chen]{wang2024msa}
Shuo Wang, Yuanhong Wang, Yanjun Peng, and Xue Chen.
\newblock Msa-net: Multi-scale feature fusion network with enhanced attention module for 3d medical image segmentation.
\newblock \emph{Computers and Electrical Engineering}, 120:\penalty0 109654, 2024.

\bibitem[Wu et~al.(2023{\natexlab{a}})Wu, Ji, Liu, Fu, Xu, Xu, and Jin]{wu2023medical}
Junde Wu, Wei Ji, Yuanpei Liu, Huazhu Fu, Min Xu, Yanwu Xu, and Yueming Jin.
\newblock Medical sam adapter: Adapting segment anything model for medical image segmentation, 2023{\natexlab{a}}.

\bibitem[Wu et~al.(2023{\natexlab{b}})Wu, Wang, Zheng, Li, Alsaadi, and Zeng]{wu2023aggn}
Peishu Wu, Zidong Wang, Baixun Zheng, Han Li, Fuad~E Alsaadi, and Nianyin Zeng.
\newblock Aggn: Attention-based glioma grading network with multi-scale feature extraction and multi-modal information fusion.
\newblock \emph{Computers in Biology and Medicine}, 152:\penalty0 106457, 2023{\natexlab{b}}.

\bibitem[Xie et~al.(2024)Xie, Zhang, and Xu]{xie2024db}
Feng Xie, Fengxiang Zhang, and Shuoyu Xu.
\newblock Db-fcn: An end-to-end dual-branch fully convolutional nucleus detection model.
\newblock \emph{Expert Systems with Applications}, 257:\penalty0 125139, 2024.

\bibitem[Xie et~al.(2020)Xie, Chen, Li, Shen, Ma, and Zheng]{xie2020instance}
Xinpeng Xie, Jiawei Chen, Yuexiang Li, Linlin Shen, Kai Ma, and Yefeng Zheng.
\newblock Instance-aware self-supervised learning for nuclei segmentation.
\newblock In \emph{MICCAI}, pages 341--350. Springer, 2020.

\bibitem[Xu et~al.(2023{\natexlab{a}})Xu, Hou, Zhang, Feng, Wang, Qiao, and Xie]{xu2023learning}
Jilan Xu, Junlin Hou, Yuejie Zhang, Rui Feng, Yi Wang, Yu Qiao, and Weidi Xie.
\newblock Learning open-vocabulary semantic segmentation models from natural language supervision.
\newblock In \emph{CVPR}, pages 2935--2944, 2023{\natexlab{a}}.

\bibitem[Xu et~al.(2022)Xu, Zhang, Wei, Lin, Cao, Hu, and Bai]{xu2022simple}
Mengde Xu, Zheng Zhang, Fangyun Wei, Yutong Lin, Yue Cao, Han Hu, and Xiang Bai.
\newblock A simple baseline for open-vocabulary semantic segmentation with pre-trained vision-language model.
\newblock In \emph{ECCV}, pages 736--753. Springer, 2022.

\bibitem[Xu et~al.(2023{\natexlab{b}})Xu, Tian, Liu, Wang, Yuan, Gu, Chen, Lukasiewicz, and Leung]{xu2023collaborative}
Zhenghua Xu, Biao Tian, Shijie Liu, Xiangtao Wang, Di Yuan, Junhua Gu, Junyang Chen, Thomas Lukasiewicz, and Victor~CM Leung.
\newblock Collaborative attention guided multi-scale feature fusion network for medical image segmentation.
\newblock \emph{IEEE Transactions on Network Science and Engineering}, 11\penalty0 (2):\penalty0 1857 -- 1871, 2023{\natexlab{b}}.

\bibitem[Yan et~al.(2020)Yan, Wang, Zhang, Luo, Xu, Xu, Zhang, Shi, Zhang, and You]{yan2020attention}
Qingsen Yan, Bo Wang, Wei Zhang, Chuan Luo, Wei Xu, Zhengqing Xu, Yanning Zhang, Qinfeng Shi, Liang Zhang, and Zheng You.
\newblock Attention-guided deep neural network with multi-scale feature fusion for liver vessel segmentation.
\newblock \emph{IEEE Journal of Biomedical and Health Informatics}, 25\penalty0 (7):\penalty0 2629--2642, 2020.

\bibitem[Yu et~al.(2023)Yu, Wang, Fu, Kou, Huang, Yang, Yang, and Gao]{yu2023techniques}
Ying Yu, Chunping Wang, Qiang Fu, Renke Kou, Fuyu Huang, Boxiong Yang, Tingting Yang, and Mingliang Gao.
\newblock Techniques and challenges of image segmentation: A review.
\newblock \emph{Electronics}, 12\penalty0 (5):\penalty0 1199, 2023.

\bibitem[Zhang et~al.(2024)Zhang, Huang, Jin, and Lu]{zhang2024vision}
Jingyi Zhang, Jiaxing Huang, Sheng Jin, and Shijian Lu.
\newblock Vision-language models for vision tasks: A survey.
\newblock \emph{IEEE Transactions on Pattern Analysis and Machine Intelligence}, 2024.

\bibitem[Zhao et~al.(2021)Zhao, He, Zhao, Huang, and Zuo]{zhao2021net}
Jing Zhao, Yong-Jun He, Si-Qi Zhao, Jin-Jie Huang, and Wang-Meng Zuo.
\newblock Al-net: Attention learning network based on multi-task learning for cervical nucleus segmentation.
\newblock \emph{IEEE Journal of Biomedical and Health Informatics}, 26\penalty0 (6):\penalty0 2693--2702, 2021.

\bibitem[Zhao et~al.(2023)Zhao, Fu, Tian, Song, and Sham]{zhao2023gsn}
Tengfei Zhao, Chong Fu, Yunjia Tian, Wei Song, and Chiu-Wing Sham.
\newblock Gsn-hvnet: A lightweight, multi-task deep learning framework for nuclei segmentation and classification.
\newblock \emph{Bioengineering}, 10\penalty0 (3):\penalty0 393, 2023.

\bibitem[Zhou et~al.(2022)Zhou, Yang, Loy, and Liu]{zhou2022learning}
Kaiyang Zhou, Jingkang Yang, Chen~Change Loy, and Ziwei Liu.
\newblock Learning to prompt for vision-language models.
\newblock \emph{International Journal of Computer Vision}, 130\penalty0 (9):\penalty0 2337--2348, 2022.

\end{thebibliography}
}


\end{document}